\documentclass[11pt]{article}
\usepackage[margin=1in]{geometry}
\usepackage{amsmath,amssymb,booktabs,graphicx,hyperref,xcolor}
\usepackage{pgfplots}\pgfplotsset{compat=1.17}
\hypersetup{colorlinks=true,linkcolor=blue,citecolor=blue,urlcolor=blue}
\usepackage{authblk}

\title{The Ignition Is Real, and It Lives at the Readout:\\
\large Latent composition, difficulty-clocked ignition, and the interface-constituted commit in a recurrent-depth reasoner}
\author[1]{Simon Lam-Muir}
\affil[1]{Prime Calibre, Australia \quad \texttt{research@primecalibre.com} \quad ORCID 0009-0001-2442-4479}
\date{July 2026}

\begin{document}
\maketitle

\begin{abstract}
We test whether the ``compositional ignition'' signature reported for latent-reasoning language models is (a) a real property of the computation, (b) an artifact of instruments or readout, or (c) inherited from human verbal training data. We grow an independent realization of a published 30M-parameter recurrent-depth reasoner from scratch under its published recipe (same configuration and seed), film its development at 8-epoch resolution (archive from epoch 328; the earliest window predates it --- see the incident records), certify its fidelity to the released reference through a pre-registered whole-signature gate, and measure resolution dynamics in two channels simultaneously --- the vocabulary readout (rank) and the hidden state ($\|\Delta h\|$) --- through estimators validated on known answers before any verdict data. 

\textbf{Result: the ignition is real, and it lives at the readout.} The rank-channel signature --- arrival time rising lawfully with problem depth (first-hit median $1\to4$ across $k=1$--$10$), sharp single-iteration resolution (runway-qualified sharpness 0.805 $\in$ [0.714, 0.867]), and arrive-and-hold stability (in-window departure $\le$ 0.07 everywhere) --- meets every pre-registered COHERENCE criterion and reproduces across two same-seed, chaotically divergent realizations of the training recipe at two curriculum stages. The resolution is an interface event at every depth. The composed intermediates were never recoverable through the tied vocabulary readout under the preregistered rank criterion (relay $= 0.00$ at every iteration, $k$, and checkpoint). A preregistered scale-invariance check rejected our earlier velocity-trough interpretation (Section~\ref{sec:commit}). What replaces it is sharper: at commitment the decision margin jumps 5.8--8.0 logits in a single iteration, exceeding the 90th percentile of near-threshold non-event steps in 96\% of cases. The signed margin's zero-crossing at that iteration is definitional and carries no evidential weight; the evidence is this conditioned magnitude. The hidden-state direction snaps in raw geometry, meeting its preregistered criterion; in the decoder's LayerNorm coordinates the arrival step falls just below our preregistered bar, so the composite decoder-coordinate claim is not confirmed (Section~\ref{sec:commit}), though decoder-coordinate trajectories descriptively show the same rapid post-commit stabilisation. The earlier paradox dissolves: the ``post-commit motion'' was decode-invisible scale; at commitment the answer margin changes abruptly and the readout-relevant geometry enters a stable decision basin, progressively converging, while subsequent raw-state displacement is predominantly radial (0.961 of squared-norm) and readout-null to a measured bound.

Three dissociations scope the finding. \emph{Cross-realization robustness:} the rank-channel signature is robust across two same-seed realizations with nondeterministically divergent training trajectories (byte-matching clocks across realizations), while state-basin geometry records the specific route and realization. \emph{Development:} the ``settled decision'' signatures are late achievements, absent when competence arrives and grown over en-route training; frozen-stage consolidation did not reproduce them in the one available comparison, which is itself confounded. \emph{Language-independence:} all of this emerged in a model trained purely on non-verbal synthetic structure, so corpus inheritance is not necessary to produce this signature in this substrate class.
\end{abstract}

\section{The questions}
\textbf{Q1 (the field's):} is workspace-like structure in latent reasoners a functional global workspace (WORKSPACE), an artifact of the output/verbalization pathway (PRINTER), or a reflection of human-language training data describing minds (MIRROR)?

\textbf{Q2 (the founding question):} is the workspace real --- does anything hold a stage --- or is it the shadow of the output commit: an evolving state plus a read-event that constitutes the ``decision'' at the interface (COHERENCE vs.\ PORT)?

\section{Substrate and instruments}
30M-parameter recurrent-depth transformer (768-d state, 4-layer recurrent block), trained on a closed world of 2{,}000 synthetic facts (200 entities $\times$ 10 relations), curriculum by hop-depth to stage 21, replicating the published recipe of the reference model. Instruments: dual-channel per-iteration capture (rank of answer and of every walk-verified intermediate; $\|\Delta h\|$ at answer position and whole-sequence; top-5 margins), padded/eval-true convention throughout, decode fast-path proven byte-identical before use. All estimators (first-hit clock, runway-qualified sharpness, windowed departure, phase-resolved $\|\Delta h\|$) were validated by exact reproduction of reference-class numbers before any verdict capture. Event detection and event-aligned state measurement are \emph{not} independent observations: a single forward pass per item hooks the final block, and both channels are deterministic functions of the same captured hidden states --- the rank channel is $\mathrm{rank}(W_U\,\mathrm{LN}(h_r))$ and the state channel is $\|h_r\|$ and $\|h_{r+1}-h_r\|$ of those same $h_r$. They are coherent by construction, which is the design intent (the axes are commensurable) and a stated dependence, not an assumed independence. All readings were pre-registered; every criterion was frozen before its data existed; every amendment is reference-side, versioned, with evidence blocks. Full methods, the predictions ledger, incident records, and the developmental film index ship in the companion repository (\url{https://github.com/primecalibre-research/ltg-replication-receipts}).

\section{Findings}
\begin{table}[t]\centering\footnotesize\setlength{\tabcolsep}{4pt}
\caption{Margin geometry at the commit, our realization, frame 4257. Rows cover all \emph{event-eligible} items: those that arrive (rank 1 at some iteration) and have a defined pre-arrival step ($r^*>0$); at $k{=}2$ a single item qualifies under the first-hit distribution, hence no CI. The \emph{signed} target margin is $\mathrm{logit(correct)}-\max_{t\neq\mathrm{correct}}\mathrm{logit}(t)$, computed exactly through the model's readout; it is negative before the commit and positive at it in 274/274 event-eligible items; the sign change at the commit is definitional (rank~1 at that iteration is equivalent to a positive signed margin), so the informative quantity is its magnitude, not the count. The \emph{null} is the within-item, same-quantity reference: $|\Delta\mathrm{margin}|$ at every iteration \emph{except} the commit step, on the same trajectory. CIs are bootstrap on the median (2000 resamples); the cluster bootstrap resamples head entities rather than items, matching the single-world generative structure.}
\label{tab:margin}
\begin{tabular}{rrrrrllrrrr}\toprule
& & \multicolumn{5}{c}{signed target margin} & \multicolumn{4}{c}{unsigned top1$-$top2} \\\cmidrule(lr){3-7}\cmidrule(lr){8-11}
$k$ & $n$ & pre & commit & $|\Delta|$ & item 95\% & cluster 95\% & $|\Delta|$ & ratio & null med. & null p90 \\\midrule
2 & 1 & -2.17 & 7.16 & 9.33 & --- & --- & 6.81 & 7 & 0.200 & 0.969 \\
3 & 39 & -1.10 & 8.54 & 9.49 & [8.76, 10.16] & [8.84, 10.19] & 8.00 & 12 & 0.150 & 0.676 \\
4 & 39 & -0.81 & 6.30 & 7.34 & [6.42, 8.46] & [6.42, 8.44] & 6.18 & 9 & 0.124 & 0.693 \\
5 & 39 & -0.59 & 6.63 & 7.05 & [6.55, 7.95] & [6.55, 8.08] & 6.60 & 11 & 0.117 & 0.586 \\
6 & 39 & -2.74 & 6.97 & 9.57 & [8.52, 10.58] & [8.57, 10.58] & 5.76 & 8 & 0.104 & 0.704 \\
7 & 29 & -0.93 & 6.84 & 7.38 & [6.71, 8.38] & [6.71, 8.38] & 6.73 & 14 & 0.078 & 0.494 \\
8 & 30 & -0.44 & 5.89 & 6.42 & [5.99, 7.26] & [5.99, 7.32] & 5.76 & 19 & 0.073 & 0.310 \\
9 & 29 & -2.31 & 6.83 & 9.45 & [7.56, 10.44] & [7.56, 10.44] & 6.06 & 14 & 0.077 & 0.420 \\
10 & 29 & -0.84 & 6.35 & 7.22 & [6.61, 8.12] & [6.61, 8.23] & 6.24 & 24 & 0.072 & 0.263 \\
\bottomrule\end{tabular}\end{table}
\begin{table}[t]\centering\small
\caption{C5.5 whole-signature fidelity, both pairings. The $\|\Delta h\|$ row fails in both and is headlined, not hidden. Verbatim from \texttt{results/c5\_5\_pairing\_a.json} and \texttt{results/c5\_5\_row2.json}.}
\label{tab:c55}
\begin{tabular}{llccccc}\toprule
pairing & T vs.\ P & clock & departure & sharpness & $\|\Delta h\|$ & all-pass \\\midrule
(a) & 3992 vs.\ 5019 & \checkmark & \checkmark & \checkmark & \textbf{fail} & \textbf{fail} \\
(b) & 4257 vs.\ 5388 & \checkmark & \checkmark & \checkmark & \textbf{fail} & \textbf{fail} \\
\bottomrule\end{tabular}\end{table}
\subsection{The substrate is certified}
The independent realization reproduces the reference's difficulty clock (first-hit medians byte-matching at every $k$; slope CI overlap at two stages), sharpness band, and hold. Fidelity is provable on every rank-channel family and was proven twice (stage-12 and stage-21 pairings).

\begin{table}[t]\centering\small
\caption{Per-$k$ grid on 340 pre-committed items (our realization, frame 4257). Sharpness is runway-qualified ($r^*\ge4$), so its $n$ is small at low $k$ by construction. Verbatim from \texttt{results/asset\_c\_grid.json} in the receipts repository.}
\label{tab:grid}
\begin{tabular}{rrrrlrrr}\toprule
$k$ & $n$ & arrived & first-hit & 95\% CI & sharpness ($n$) & departure & crossed floor \\\midrule
1 & 20 & 20 & 1.0 & [1.0, 1.0] & --- & 0.0 & 1.0 \\
2 & 40 & 39 & 1 & [1.0, 1.0] & --- & 0.0 & 0.975 \\
3 & 40 & 39 & 2 & [2.0, 2.0] & --- & 0.0 & 0.975 \\
4 & 40 & 39 & 2 & [2.0, 2.0] & --- & 0.0 & 0.975 \\
5 & 40 & 39 & 2 & [2.0, 2.0] & --- & 0.026 & 0.975 \\
6 & 40 & 39 & 3 & [3.0, 3.0] & 0.59 (1) & 0.0 & 0.975 \\
7 & 30 & 30 & 3.0 & [3.0, 3.0] & --- & 0.067 & 1.0 \\
8 & 30 & 30 & 3.0 & [3.0, 3.0] & 0.805 (2) & 0.0 & 1.0 \\
9 & 30 & 30 & 4.0 & [4.0, 4.0] & 0.807 (27) & 0.033 & 1.0 \\
10 & 30 & 29 & 4 & [4.0, 4.0] & 0.791 (29) & 0.0 & 0.967 \\
\bottomrule\end{tabular}\end{table}
\begin{figure}[t]\centering
\begin{tikzpicture}\begin{axis}[width=.47\textwidth,height=4.6cm,xlabel={iterations from commit},ylabel={signed target margin},legend style={font=\tiny,at={(0.03,0.97)},anchor=north west},xtick={-3,-1,1,3,5,7},grid=major,grid style={gray!20}]
\addplot+[mark=*,mark size=1pt] coordinates {(-1,-0.9283) (0,7.563) (1,10.95) (2,10.97) (3,10.72) (4,10.47) (5,10.31) (6,10.25) (7,10.11) (8,9.999)};
\addlegendentry{$k$ 2--4}
\addplot+[mark=*,mark size=1pt] coordinates {(-3,-0.4807) (-2,-0.4738) (-1,-0.7586) (0,6.78) (1,9.933) (2,10.16) (3,9.99) (4,10.01) (5,9.828) (6,9.731) (7,9.562) (8,9.482)};
\addlegendentry{$k$ 5--7}
\addplot+[mark=*,mark size=1pt] coordinates {(-3,-0.3927) (-2,-0.4376) (-1,-0.7703) (0,6.352) (1,9.463) (2,9.912) (3,9.874) (4,9.759) (5,9.582) (6,9.452) (7,9.368) (8,9.309)};
\addlegendentry{$k$ 8--10}
\draw[dashed,gray] (axis cs:0,\pgfkeysvalueof{/pgfplots/ymin}) -- (axis cs:0,\pgfkeysvalueof{/pgfplots/ymax});
\end{axis}\end{tikzpicture}\hfill
\begin{tikzpicture}\begin{axis}[width=.47\textwidth,height=4.6cm,xlabel={iterations from commit},ylabel={answer rank},ymode=log,legend style={font=\tiny},xtick={-3,-1,1,3,5,7},grid=major,grid style={gray!20}]
\addplot+[mark=*,mark size=1pt] coordinates {(-1,47) (0,1) (1,1) (2,1) (3,1) (4,1) (5,1) (6,1) (7,1) (8,1)};
\addplot+[mark=*,mark size=1pt] coordinates {(-3,79) (-2,90.5) (-1,55) (0,1) (1,1) (2,1) (3,1) (4,1) (5,1) (6,1) (7,1) (8,1)};
\addplot+[mark=*,mark size=1pt] coordinates {(-3,93) (-2,87) (-1,67.5) (0,1) (1,1) (2,1) (3,1) (4,1) (5,1) (6,1) (7,1) (8,1)};
\end{axis}\end{tikzpicture}
\caption{Event-aligned readout. Left: signed target margin (correct $-$ best incorrect) crosses zero at the commit in every event-eligible item. Right: answer rank (log scale). Medians over items; offset 0 is the commit iteration.}\label{fig:readout}\end{figure}
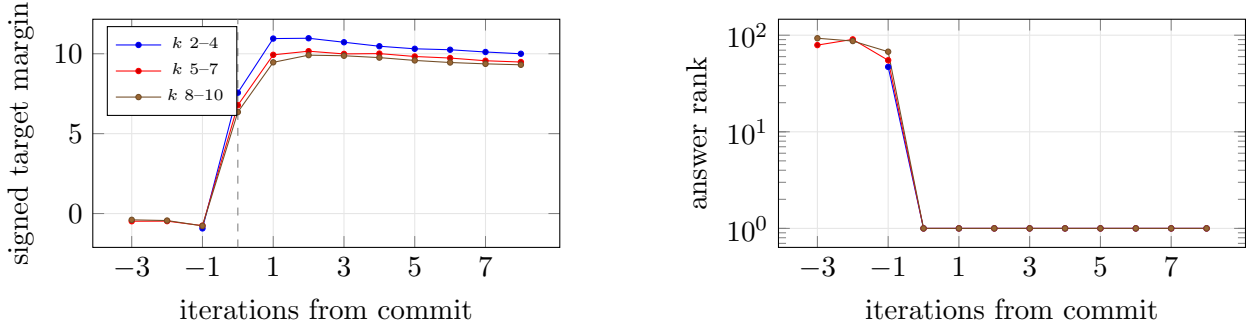

\subsection{The ignition signature is real structure}
Per-$k$ table on 340 pre-committed novel items: clock rises $1\to4$ across $k=1$--$10$; sharpness 0.805 in-band on runway-clean cells; departure holds; crossed-differential floor 0.97--1.00 (the signal is query-specific everywhere --- not decode-side salience, not corpus statistics). All three COHERENCE conjuncts met at frozen thresholds; claimed for the reference class on two-stage cross-realization reproduction.

\paragraph{Directional metrics (definitions).} For consecutive states $x_r,x_{r+1}$ the angular step is $\theta_r=\arccos\!\big(\langle x_r,x_{r+1}\rangle/(\|x_r\|\,\|x_{r+1}\|)\big)$, and the directional \emph{displacement} is $1-\cos\theta_r$. Reported ratios are of this displacement under the phase decomposition used throughout (PRE: steps before first-hit; EVENT: the arrival step; POST: steps from first-hit onward), aggregated as means over pooled steps. These are displacement ratios and may exceed 1; they are not cosine similarities.

\begin{figure}[t]\centering
\begin{tikzpicture}\begin{axis}[width=.47\textwidth,height=4.6cm,xlabel={iterations from commit},ylabel={$\|h\|$},xtick={-3,-1,1,3,5,7},grid=major,grid style={gray!20}]
\addplot+[mark=*,mark size=1pt] coordinates {(-3,103.1) (-2,97.56) (-1,74.03) (0,47.99) (1,123.1) (2,258.4) (3,391) (4,522) (5,653.7) (6,778.4) (7,906.1) (8,1031)};
\end{axis}\end{tikzpicture}\hfill
\begin{tikzpicture}\begin{axis}[width=.47\textwidth,height=4.6cm,xlabel={iterations from commit},ylabel={angular step (deg)},legend style={font=\tiny},xtick={-3,-1,1,3,5,7},grid=major,grid style={gray!20}]
\addplot[fill=blue!12,draw=none,forget plot] coordinates {(-3,9.534) (-2,43.05) (-1,88.71) (0,62.18) (1,22.69) (2,7.463) (3,4.328) (4,3.034) (5,2.319) (6,1.868) (7,1.553) (8,1.323) (8,0.9706) (7,1.156) (6,1.404) (5,1.76) (4,2.297) (3,3.306) (2,5.361) (1,11.99) (0,54.83) (-1,76.63) (-2,13.96) (-3,8.514)}--cycle;
\addplot[fill=red!12,draw=none,forget plot] coordinates {(-3,20.09) (-2,59.49) (-1,81.92) (0,56.03) (1,21.56) (2,7.021) (3,4.074) (4,2.816) (5,2.198) (6,1.786) (7,1.49) (8,1.303) (8,0.9946) (7,1.176) (6,1.418) (5,1.776) (4,2.322) (3,3.325) (2,5.534) (1,13.22) (0,49.87) (-1,73.5) (-2,25.82) (-3,18.58)}--cycle;
\addplot+[mark=*,mark size=1pt] coordinates {(-3,8.912) (-2,17.58) (-1,83.58) (0,58.68) (1,16.03) (2,6.369) (3,3.832) (4,2.692) (5,2.046) (6,1.638) (7,1.357) (8,1.157)};\addlegendentry{raw}
\addplot+[mark=square*,mark size=1pt] coordinates {(-3,19.15) (-2,32.42) (-1,79.19) (0,52.86) (1,16.25) (2,6.178) (3,3.671) (4,2.57) (5,1.969) (6,1.6) (7,1.338) (8,1.152)};\addlegendentry{LN}
\end{axis}\end{tikzpicture}
\caption{Event-aligned state geometry. Left: $\|h\|$ contracts to a minimum at the commit and then grows near-linearly (exploratory; see Section~\ref{sec:limits}). Right: angular step in raw coordinates and in the decoder's LayerNorm coordinates --- the post-commit freeze appears in both; shaded bands are interquartile ranges across items. Medians over items.}\label{fig:state}\end{figure}
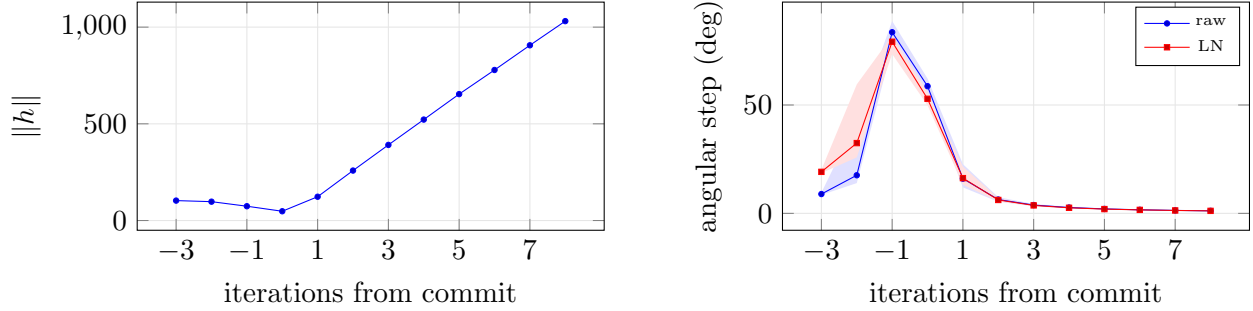

\subsection{The commit is a directional event; the velocity trough is withdrawn}\label{sec:commit}
The measurement history in full: a frozen EVENT/PRE ratio read 1.15 (interface, mild high-$k$ drift); a pre-flagged follow-on found a denominator contamination, and the clean ratio suggested deep-$k$ substance-spikes (2.5--4$\times$ the descent); the pre-registered four-cell adjudicator then demoted the spike reading --- the arrival step never clears the phase-neutral velocity floor at any $k$, and the percentile statistic located it at the $\sim$1st percentile of plateau motion. That reading is now \textbf{withdrawn}. Pre-registered normalization controls (raw, RMS-scaled, norm-relative, and directional re-representation, at answer position and whole sequence) show the trough survives only in the magnitude-carrying families and \emph{inverts} under norm-relative and directional ones (post/pre 0.40 and 0.06 against raw's 2.00). The trough tracked the state's magnitude arc, not a computational quieting. What survives, and is sharper, is directional. In raw state geometry the readout-relevant direction snaps at the commit ((1$-$cos) displacement EVENT/PRE 1.62) and then progressively freezes (POST/PRE 0.059), while the magnitude grows in a channel the readout cannot see (Section~\ref{sec:limits}). Repeating the analysis in the decoder's own coordinates $z_r=\mathrm{LN}(h_r)$ --- the coordinates the logits are a function of --- the arrival step attenuates to 1.46, missing the pre-registered 1.5 bar by 0.04, while the freeze reproduces (POST/PRE 0.051; LN angular steps $52.9^\circ$ at the commit decaying to $1.2^\circ$ by the eighth iteration, $n{=}274$). Decomposing each post-commit step into components along and orthogonal to the pre-step state, the radial share is 0.961 of squared-norm (IQR [0.937, 0.974], $n{=}4{,}826$ steps); moving the state along that radial component shifts the logits by at most $5.7\times10^{-6}$ (90th percentile) against a tangential effect of 0.18, and a separate scaling test bounds pure rescaling at $7\times10^{-5}$ with zero rank changes. The attenuation locates its own mechanism: LayerNorm's mean-subtraction and affine transform absorb part of the arrival-step displacement. Per the registered consequence the snap is therefore reported as a \emph{raw-geometry} finding; the composite pre-registered claim (snap \emph{and} freeze in decoder coordinates) is \textbf{not} confirmed, and the freeze figure above is offered as description of the table, not as a separately registered result. A decoder-coordinate snap claim would require a fresh registered read on independent data (the 3992 frame or the reference class), which we have not run. Event-counts rise with $k$ (slope 0.353, CI [0.088, 0.855]) without hop-ordered relay --- a third independent silent-walk signature. Four-cell occupancy on our realization matches the reference class qualitatively (ATTRACTOR-dominated: 0.672 vs.\ 0.739).

\subsection{Cross-realization robustness vs.\ route-dependence}
Rank-channel: same-seed cross-realisation robust, established across two same-seed realizations (robustness to nondeterministic trajectory divergence); recipe-level invariance across independent seeds is untested and registered as future work. State-basin: realization- and route-dependent --- the run's property. Both fidelity rows failed the state channel \emph{as the finding}: no faithful copy can share another realization's path, so state geometry cannot serve as a cross-model fidelity criterion; it was demoted to a caveated realization reading (pre-drawn branch, ratified) while remaining the within-realization co-primary.

\subsection{Commit signatures emerge after competence}
The developmental film shows one expensive reorganization (the first composition, $\sim$2{,}808 epochs), then a depth-general operation transferring and extrapolating; capabilities assembled piecewise --- relational content first, transport dynamics and fast readout co-forming later; basin depth absent at competence and grown over en-route training; the consolidation twin \emph{suggests} dwelling at a frozen stage does not reproduce it (the twin's clock sharpens while its basin sinks), though that comparison's interpretation is limited by the calibration shift the sink itself induces (Section~\ref{sec:limits}). An unconfounded test requires an intervention we have not run. The signatures of settled decision are acquired developmental features in this training trajectory rather than features already present at competence onset.

\section{The verdict}
\textbf{MIRROR --- not necessary to produce this signature in this substrate class.} The full ignition signature emerged in a model that has never seen a word of human language. Workspace-like ignition does not require verbal training data. (The global-workspace reading of verbalizable representations in frontier models is due to Gurnee et al.~\cite{workspace}.) (Whether frontier models' verbalizable workspace additionally inherits corpus structure remains open; the existence of the signature no longer needs that hypothesis.)

\textbf{WORKSPACE (strong form) --- not evidenced in the channels measured here.} The walk is silent, the intermediates never surface through the tied vocabulary readout under the preregistered criterion, and ``settled'' is a property of the readout class: the state's readout-relevant geometry enters a stable basin at commitment and rapidly converges thereafter, while its readout-null magnitude keeps growing. On these instruments, no staged broadcast is detected: the readout class is stable while the state moves --- the stage, as measured, is the reading. (A moving state can still maintain information --- the decode-basin is precisely such maintenance; what is absent is evidence of broadcast or staged intermediates.)

\textbf{PRINTER (dismissive form) --- insufficient.} The interface event is lawful, reliable and difficulty-lawful: clocked by computational depth with a clean linear law, difficulty-graded in sharpness, held once crossed, reproduced across two same-seed realizations. Whether the event is \emph{used} is untested here: our captures run fixed iteration ladders, so we have no halting analysis; the registered early-exit intervention (forcing readout immediately before versus after the event) remains the path to any functional claim. The continuous readout geometry moves with the rank: at the commit the decision margin jumps 5.8--8.0 logits in a single iteration, 7--24$\times$ the 90th percentile of non-event steps (Table~\ref{tab:margin}). Because the event is by definition the first rank-1 iteration, the signed margin's zero-crossing there is definitional and carries no evidential weight; the evidence is the conditioned magnitude. A tightly-matched per-$k$ null could not be constructed: conditioning on pre-step margin is near-collinear with the event definition itself (1.4\% of non-event steps occupy the event's margin region, and a fourfold bin widening leaves the match rate unchanged) --- a limitation we report rather than relax. Against a regression baseline fitted on all 5{,}040 non-event steps ($|\Delta| \sim$ pre-margin $+$ iteration $+ k$), 268/274 event residuals exceed the non-event 90th percentile (median residual 6.33 against 0.36). Against the identifiable near-threshold null --- the 71 non-event steps lying inside the event's own pre-margin interquartile range, i.e.\ approaches to the boundary that did not cross --- the arrival step's median $|\Delta|$ is 7.96 against an in-band 90th percentile of 3.15, with 96.4\% of events above it: the transition is not the generic behavior of margins near the boundary. These results argue strongly against the transition being explained by rank discreteness alone. The clock governs \emph{when} the transition occurs; its magnitude is depth-flat while the non-event background quiets with depth.

\textbf{The verdict: the workspace-signature is real, reliable and difficulty-lawful, and constituted at what we name a \emph{commit surface}} --- the interface at which a silently-composed computation becomes a discrete, stable, difficulty-clocked decision --- \textbf{and the commit is where the readout-relevant geometry enters its stable basin, at every depth.} The commit signature, wherever we measured it, was interface-constituted --- and the substance's role is cleaner than coupling: its readout-relevant geometry enters a stable basin at commitment and rapidly converges thereafter, while its readout-null magnitude grows. In the founding question's terms: COHERENCE in the rank channel; in the substance, no externalized port and no staged broadcast --- an interface-constituted commit. We note the registered PORT hypothesis --- an externalized sequence-position write reread by downstream computation --- is structurally absent in this architecture; the commit we observe is a readout-interface event, not a recurrent port, and we name it accordingly. The dichotomy was the error across channels, and the apparent depth-exception dissolved under the pre-registered adjudicator. The channel distinction, the margin-jump characterization of the commit, and the radial/angular dissolution are this work's contribution.

\section{Why these claims survive hostile reading}
Every criterion predates its data (append-only pre-registration; the receipts repository documents each freeze). One of us proposed, before any instrument existed to test it, that the workspace may be the shadow of the output commit; a pre-registered, timestamped predictions ledger records the hypothesis, its odds, and its confirmation. Both fidelity-gate state-channel failures are headlined, not hidden. The instrument-convention incident (an artifact that manufactured a coherent wrong law, caught by reuse-verification) is published as a methods exhibit, as are: the runway qualifier, the phase-neutral floor recalibration, the arrival-geometry check, and the predictions ledger (complete, with every resolution appended under criteria frozen before the relevant results; provenance of pre-commitment, not evidence for the claims; full ledger in the repository). Recent entries: a two-day-old headline (``depth-dependent locus'') demoted pre-release by the pre-registered four-cell adjudicator; a sign-wrong verbal hedge caught by requiring the percentile; and this paper's own velocity-trough headline, withdrawn when pre-registered normalization controls run during pre-submission review showed it to be coordinate-dependent (ledger entries L26--L27b). The programme's evidential standard is demonstrated by the claims it rejected under criteria frozen before the relevant results were observed.

\section{Related work}
The reference paper itself established that first-hit iteration tracks hop complexity in-distribution (their Figs.\ 8--9)~\cite{ltg}; our contributions on this axis are its independent reproduction from a separately grown realization, the margin-jump characterization of the commit, the directional commit --- a raw-geometry snap with a freeze that reproduces in decoder coordinates --- and its readout-null radial complement, and the developmental emergence of all of these. 
Lu et al.\ found no structured latent pathway in depth-recurrent rank trajectories; our silent walk corroborates in this regime, and our positive structure (the clocked ignition, the directional commit) is what rank-only instruments cannot see. The reference paper's own training curve independently shows the grind-then-race dynamic our stage-cost law measures. DiscoLoop reports decodable bridge entities in a two-hop regime --- intermediate decodability is training-regime-dependent (present there, absent here and in Huginn), itself informative. Width-based latent models (Coconut/CODI) show partially decodable traces --- a class contrast. The halting-gates line studies when looped models stop; we ask where the decision is constituted. A supervision-side result argues loss controls only readout-exposed variables --- a training-time cousin of our thesis. Full citations in the bibliography.

\section{Limitations and future work}\label{sec:limits}
The near-threshold null supporting the margin claim comprises 71 control steps, all from $k{=}6$--10: items at lower $k$ commit within one or two iterations and so never spend a non-event step near the threshold, contributing no controls. The comparison is therefore 274 events against 71 controls drawn from 1.4\% of the non-event population. Statistical intervals are conditional on the sampled synthetic world; generalisation across world realisations requires retraining or evaluating on independently generated fact systems. 
The snap is answer-position-specific: measured over the whole padded sequence the directional event does not appear ((1$-$cos) displacement event/pre 0.87 vs 1.62 at the answer position) and raw $\|\Delta h\|$ post/pre inverts (0.85 vs 1.99), as expected when the commit position is averaged against padding; the post-commit directional freeze appears in both ((1$-$cos) displacement post/pre 0.06 and 0.02). The $\alpha$-scaling and radial/angular tests use the model's real readout; LayerNorm is scale-invariant only in the $\epsilon\to0$ limit, so the radial channel is readout-null up to a term of relative order $\epsilon/(\alpha^2\mathrm{var}(h)+\epsilon)$, measured at $\le7\times10^{-5}$ absolute with zero rank changes; at that magnitude the $\epsilon$ term and float32 rounding are not separable. A magnitude arc accompanies the commit --- $\|h\|$ contracts to a minimum at the commit iteration and then grows near-linearly --- which explains the withdrawn trough entirely; this observation is \textbf{exploratory}, discovered post-hoc during capture verification rather than pre-registered, and its early-iteration-transient confound was tested and refuted ($\arg\min\|h\|$ tracks first-hit across $f_h$ groups, 82\% exact alignment). 

One substrate class (30M recurrent-depth), one task family (pure-depth composition; interference/binding deliberately excluded). Rank-channel claims: reference class. State-channel readings: our realization, caveated (state geometry is realization-dependent under fixed seed --- itself a new finding). The twin four-cell comparison is reported as confounded (the basin-sink moves the instrument's own calibration distribution); the sink stands on three direct measures. Future work, promised: (1) the capacity-wall expedition, underway --- the first filmed approach to a reasoner's depth limit; (2) the language-diet arm --- does verbal training move the commit surface?; (3) the scale question --- does the commit-surface geometry hold in frontier models? (requires partners).

\section*{Contributions}
Simon Lam-Muir (Prime Calibre): founding hypothesis, programme direction, all ratification decisions, hands-on execution. Analysis design, review, and drafting in collaboration with Claude Fable 5 (Anthropic; high reasoning effort). Autonomous build and execution by a delegated coding agent running Claude Opus 4.8 (Anthropic; high reasoning effort). Curated role transcripts (symbolic-arc sessions), the predictions ledger, and every incident record ship in the receipts repository: \url{https://github.com/primecalibre-research/ltg-replication-receipts}.

\section*{One-paragraph version}
We grew an independent copy of a published latent reasoner, filmed its development, certified its fidelity through gates frozen before the data existed, and asked --- in two channels at once --- what happens when it silently decides. The answer: a real, difficulty-clocked, holding ignition that reproduces across two same-seed realizations of the recipe with divergent training trajectories --- occurring at the readout at every depth, where the decoder's view of the state snaps and then rapidly stabilises while its magnitude keeps growing unseen --- over a process that composes silently, never surfaces its intermediates through the tied vocabulary readout under the preregistered criterion, and continues moving predominantly in a radial channel the decoder barely sees. The ignition is real. It is grown late, by the whole route. And it is the shadow of the commit --- which is what one of us said it might be before we had any instrument that could check.

\end{document}